\documentclass{article}

\usepackage{PRIMEarxiv}
\usepackage{float}
\usepackage[utf8]{inputenc} 
\usepackage[T1]{fontenc}    
\usepackage[colorlinks=true]{hyperref}
\usepackage{url}            
\usepackage{booktabs}       
\usepackage{amsfonts}       
\usepackage{nicefrac}       
\usepackage{microtype}      
\usepackage{lipsum}
\usepackage{fancyhdr}       
\usepackage{graphicx}       
\usepackage{xcolor}
\usepackage{graphicx}
\usepackage{amsmath}
\usepackage{amssymb}
\usepackage{booktabs}
\usepackage{multirow}
\usepackage{tabularray}
\usepackage{array}
\usepackage{xspace}

\makeatletter
\DeclareRobustCommand\onedot{\futurelet\@let@token\@onedot}
\def\@onedot{\ifx\@let@token.\else.\null\fi\xspace}

\def\ie{\emph{i.e}\onedot}

\makeatother

\title{RxRx1: A Dataset for Evaluating Experimental Batch Correction Methods

}

\author{
  Maciej Sypetkowski$^1$\\
   \And
  Morteza Rezanejad$^1$\\
    \And
  Saber Saberian$^1$\\
   \And
  Oren Kraus$^1$\\
   \And
  John Urbanik$^1$\\
   \And
  James Taylor$^1$\\
   \And
  Ben Mabey$^1$\\
   \And
  Mason Victors$^1$\\
   \And
  Jason Yosinski$^2$\\
   \And
  Alborz Rezazadeh Sereshkeh$^1$\\
   \And
  Imran Haque$^1$\\
  \And 
  Berton Earnshaw $^{1,}$\thanks{Corresponding author: Berton Earnshaw (\url{berton.earnshaw@recursion.com}). \newline  $^1$ These authors contributed to this article during their employment with Recursion.\newline $^2$ Jason Yosinksi contributed to this article as a machine learning advisor to Recursion.
} \\
}

\begin{document}
\maketitle

\begin{abstract}
High-throughput screening techniques are commonly used to obtain large quantities of data in many fields of biology. It is well known that artifacts arising from variability in the technical execution of different experimental batches within such screens confound these observations and can lead to invalid biological conclusions. It is therefore necessary to account for these \emph{batch effects} when analyzing outcomes. In this paper we describe \emph{RxRx1}, a biological dataset designed specifically for the systematic study of batch effect correction methods. The dataset consists of 125,510 high-resolution fluorescence microscopy images of human cells under 1,138 genetic perturbations in 51 experimental batches across 4 cell types. Visual inspection of the images alone clearly demonstrates significant batch effects. We propose a classification task designed to evaluate the effectiveness of experimental batch correction methods on these images and examine the performance of a number of correction methods on this task. Our goal in releasing RxRx1 is to encourage the development of effective experimental batch correction methods that generalize well to unseen experimental batches. The dataset can be downloaded at \url{https://rxrx.ai}.\\
\end{abstract}


\section{Introduction}

\begin{figure}[!ht]
    \centering
    \begin{tabular}{c@{\hskip 5pt}c@{\hskip 5pt}c}
    \includegraphics[height = 0.175\textwidth]{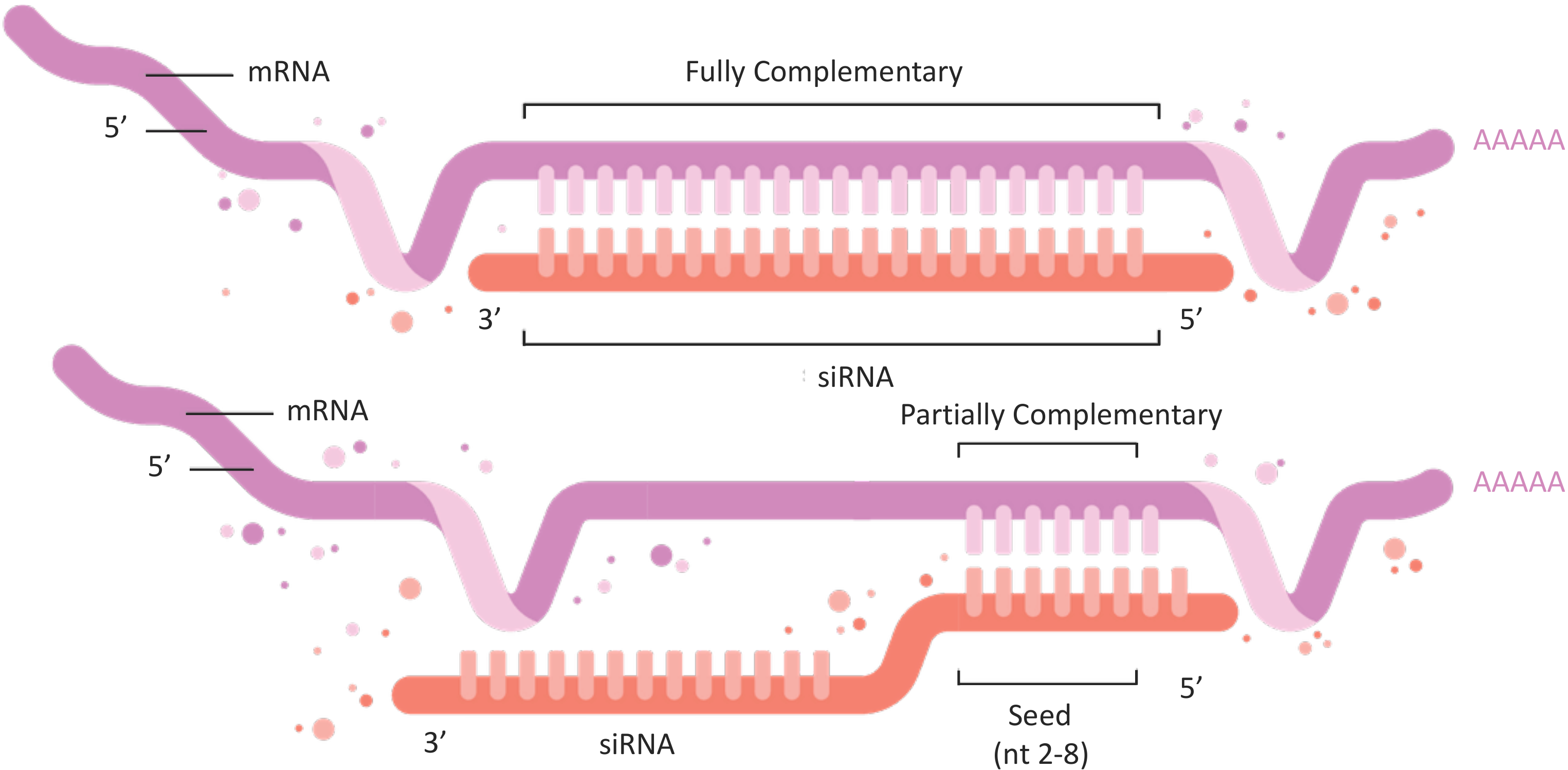}& 
    \includegraphics[height = 0.175\textwidth]{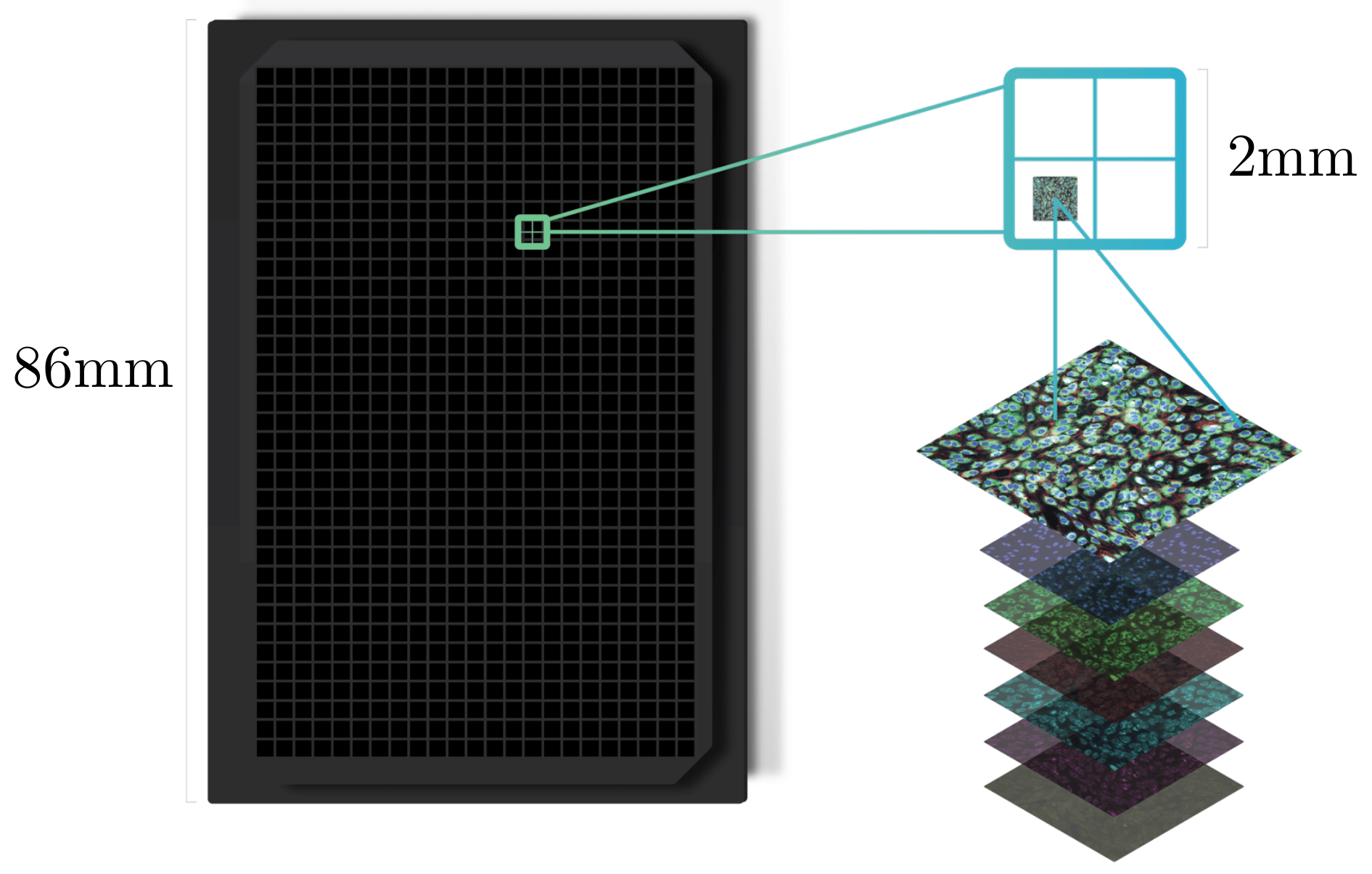} & 
    \includegraphics[height = 0.175\textwidth]{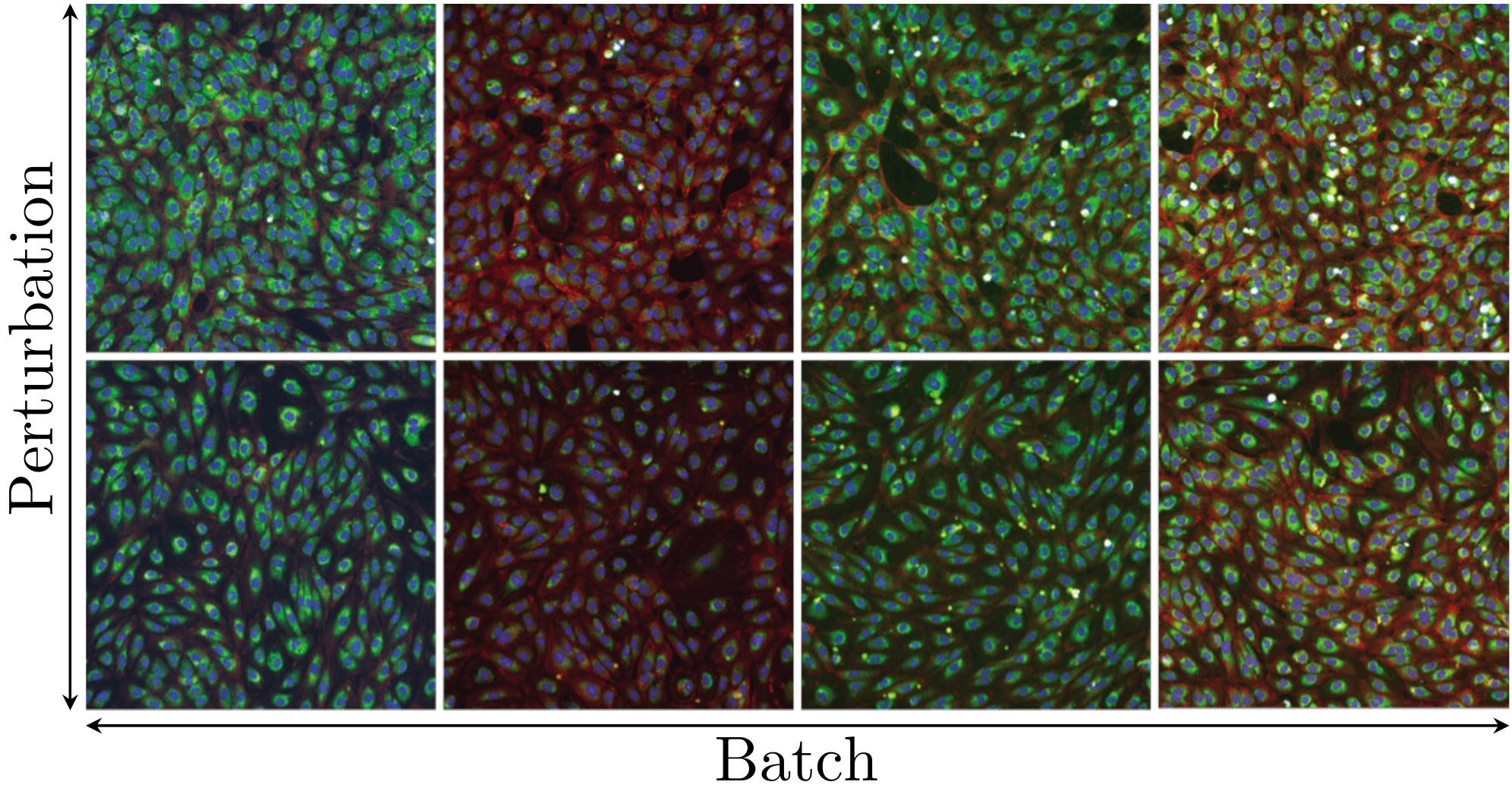}\\
    \textbf{(a)} & \textbf{(b)} & \textbf{(c)}\\
    \end{tabular}
    \caption{\textbf{(a)} Top: Depiction of full-complementarity of an siRNA to an mRNA to knockdown a particular target gene. Bottom: Depiction of partial-complementarity in the seed-region of an siRNA, leading to partial knockdown of hundreds of additional genes. \textbf{(b)} Schematic of a 384-well plate demonstrating imaging sites and 6-channel images. The experiments in this dataset were run in such plates. RxRx1 contains two 6-channel images from different sites per well.  \textbf{(c)}  Images of two different genetic conditions (rows) in HUVEC cells across four experimental batches (columns). Notice the visual similarity of images from the same batch.\\}
    \label{fig:main_fig}
\end{figure}

High-throughput screening is commonly used in many biological fields, including genetics 
\cite{echeverri2006high,zhou2014high} and drug discovery \cite{broach1996high,macarron2011impact,swinney2011were,boutros2015microscopy}. Such screens are capable of generating large amounts of data that, when coupled with modern machine learning methods, could help answer fundamental questions in biology and solve the problem of rising costs in drug discovery, which are now estimated to be over $2$ billion per approved drug \cite{scannell2012diagnosing,dimasi2016innovation}. However, creating large volumes of biological data necessarily requires the data to be generated in multiple experimental batches, or groups of experiments executed at different times under similar conditions. Even when experiments are carefully designed to control for technical variables such as temperature, humidity, and reagent concentration, the measurements taken from these screens are confounded by artifacts that arise from differences in the technical execution of each batch. Figure \ref{fig:main_fig}c demonstrates the complexity of identifying relevant biological variation and separating it from technical noise caused by these so-called \emph{batch effects}. Even when experiments are designed to control for technical variables such as temperature, humidity, and reagent concentration, batch effects unavoidably enter into the data.
Batch effects can alter factors of variation within the images that are irrelevant to the biological variables under study but are unfortunately often correlated with them. It is therefore necessary to correct for such effects before drawing any biological conclusions \cite{leek2010tackling,parker2012practical,soneson2014batch,nygaard2016methods}. Indeed, many computational methods have been designed for correcting such effects 
\cite{korsunsky2019fast, hie2019efficient, lopez2018deep, li2020deep, Lotfollahi2020, haghverdi2018batch, goh2017batch, shaham2017removal}.

In this paper, we describe the \emph{RxRx1} dataset, an image dataset systematically designed to study batch effect correction methods. The dataset consists of 125,510 6-channel fluorescence microscopy images of human cells under 1,108 different genetic perturbations (plus 30 positive control perturbations) across 51 experimental batches and 4 cell types. We propose an invariant risk minimization task \cite{arjovsky2019invariant} to gauge the effectiveness of batch effect correction methods, namely learning to classify the genetic perturbation present in each image in a set of experimental batches held out from a training set. In order for a classifier to perform well on this task, it must be able to robustly identify the discriminative morphological features associated with each genetic perturbation against a background of the latent technical variations associated with each held-out experimental batch.

In the present article, we make three main contributions:
\begin{enumerate}
\item We present a dataset (46GB, 125,510
images, 1,139 classes including one EMPTY class) for testing experimental batch effect correction, comparable in size to reference datasets such as ImageNet \cite{5206848} (155 GB, 1.2M images, 1000 classes) and other biological datasets like BBBC017 (56 GB, 64.5K images, 4903 classes) \cite{ljosa2012annotated}.
\item We introduce a specific task for evaluating the effectiveness of batch effect correction methods, accompanied by three evaluation metrics enabling users of this dataset to assess their developed methods. 
\item We demonstrate the use of a standard convolutional classifier architecture as a backbone for the task of experimental batch correction and analyze the performance of variations of this model on such task.
\end{enumerate}

This dataset and task will be of interest to the community of researchers applying machine learning methods to complex biological datasets, especially those working with image-based high-content phenotypic screens 
\cite{angermueller2016deep,kraus2016classifying,caicedo2017data,kraus2017automated,ando2017improving,chen2018rise}. In addition, we believe RxRx1 will be of interest to the larger community of machine learning researchers working in the areas of domain adaptation, transfer learning, and few-shot learning.

\begin{figure*}[!b]
    \centering
    \includegraphics[width = 1.015\textwidth]{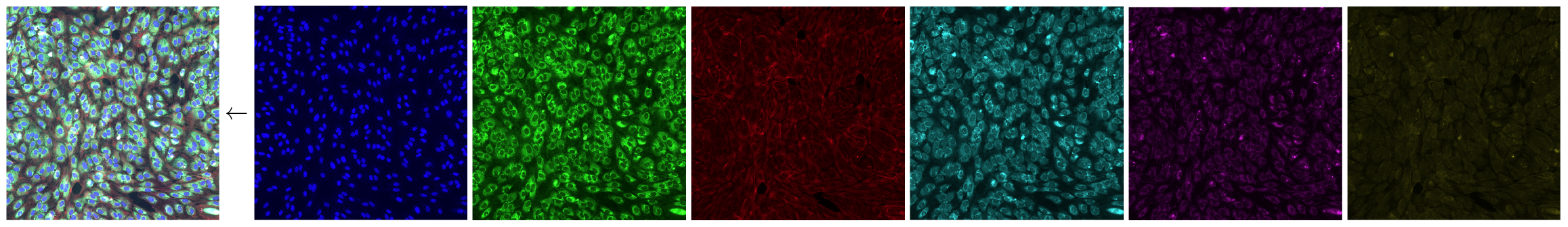}
    \caption{6-channel faux-colored composite image of HUVEC cells (left) and individual channels (right) stained with Hoechst 33342 (channel 1, blue), Alexa Fluor 488 Concanavalin A (channel 2, green), Alexa Fluor 568 Phalloidin (channel 3, red), Syto14 (channel 4, cyan), MitoTracker Deep Red FM (channel 5, magenta), and Alexa Fluor 555 Agglutinin (channel 6, yellow). The similarity in content between some channels is due in part to the spectral overlap between the fluorescent stains used in those channels.
    }

    \label{fig:fig_6_chann}
\end{figure*}

\begin{figure}[!t]
    \centering
    \includegraphics[width = 0.5\textwidth]{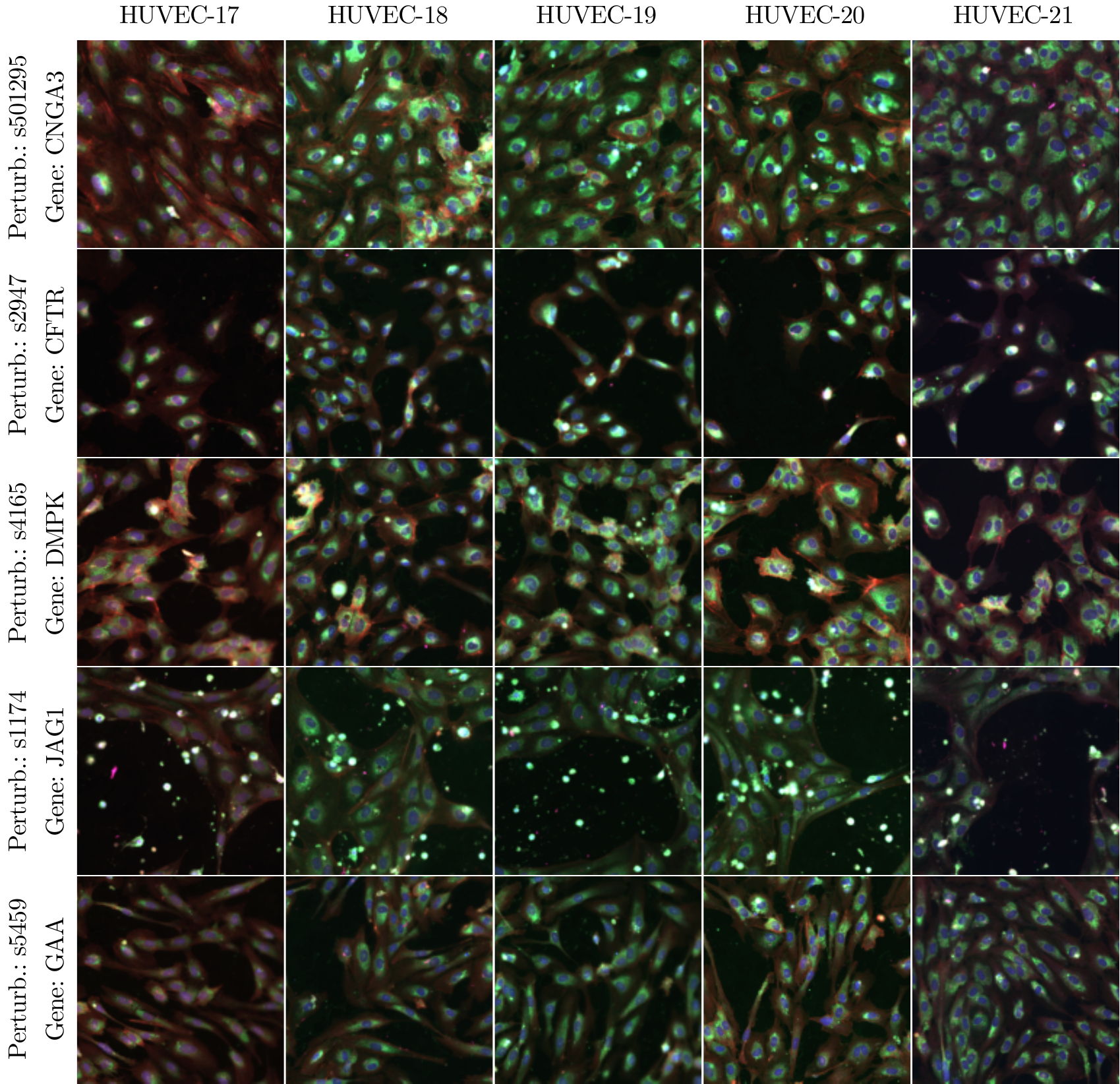}
    \caption{Images of 5 siRNA phenotypes in HUVEC cells across 5 experimental batches. Each siRNA causes changes in the visual properties of cell populations, including morphology, count, and spatial distribution.}
    \label{fig:sirna-batch-image-matrix}
\end{figure}

\section{Dataset}
\begin{figure}[!t]
    \centering
    \includegraphics[width = 0.7\textwidth]{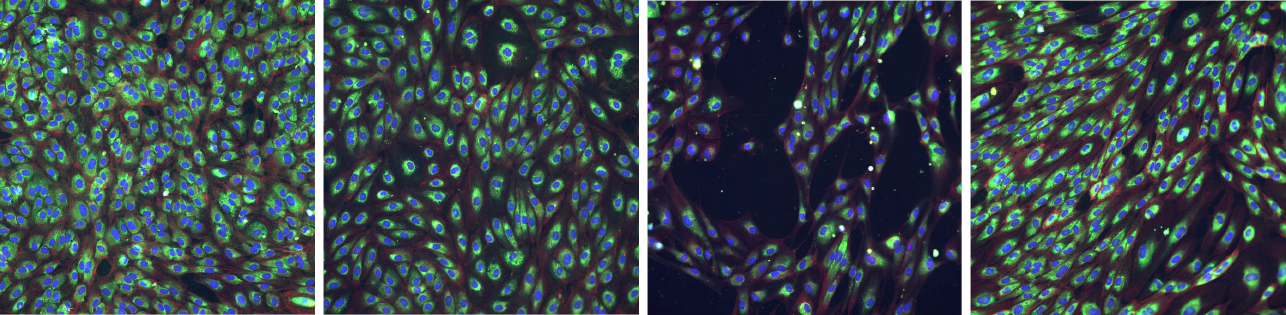}\\\vspace{0.6mm}
    \includegraphics[width = 0.7\textwidth]{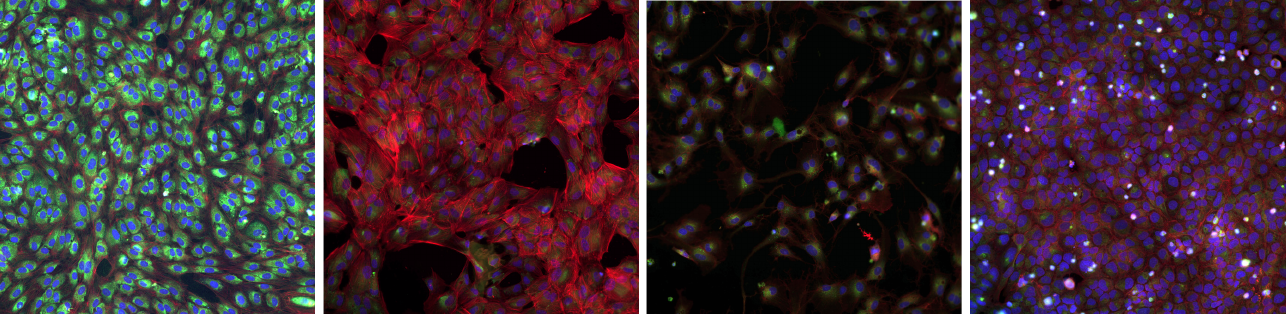}
    \caption{Top row: Images of HUVEC cells under four different siRNA perturbations, all from the same plate. Bottom row: Images of cells under the same siRNA perturbation in four cell types: HUVEC, RPE, HepG2, and U2OS.}
    \label{fig:siRNApheno}
\end{figure}

All images in RxRx1 were generated using Recursion's  high-throughput screening platform \footnote{\url{https://recursion.com}}. The dataset is comprised of fluorescence microscopy images of human cells of four different types:
\begin{itemize}
    \item HUVEC: primary endothelial cells derived from the umbilical vein \cite{davis2007isolation}.
    \item RPE: epithelial cells from the outermost layer of the retina \cite{yang2021functions}. 
    \item HepG2: nontumorigenic cells with high proliferation rates and an epithelial-like morphology important for hepatic functions \cite{donato2015culture}. 
    \item U2OS: immortalized epithelial cells derived in 1964 from an osteosarcoma patient \cite{niforou2008proteome}.
\end{itemize}
These were acquired using a proprietary implementation of the Cell Painting imaging protocol \cite{bray2016cell}. In Figure \ref{fig:fig_6_chann}, we show an example image. Each channel corresponds to a fluorescent dye used to stain one of six different targeted cellular components, namely nuclei, endoplasmic reticuli, actin, nucleoli, mitochondria, and Golgi.
The images themselves are the result of executing the same experimental design in 51 different experimental batches, with each execution separated by at least a week from all others. 
The experiment design consists of four 384-well plates (see Figure \ref{fig:main_fig}b), where each well contains an isolated population of cells. The wells are laid out on each plate in a $16\times24$ grid, but only the wells in the inner $14\times22$ grid are used since the outer wells are most susceptible to environmental factors. In each well, cell populations are genetically perturbed with small interfering ribonucleic acid, or siRNA, at a fixed concentration. Each siRNA is designed to knockdown a single target gene via the RNA interference pathway, reducing the expression of that gene \cite{tuschl2001rna}. In addition, siRNAs are known to have significant but consistent off-target effects via the microRNA pathway, creating partial knockdown of many other genes as well (see Figure \ref{fig:main_fig}a). Each siRNA, therefore, perturbs cellular function in a way that can impact visible properties of the cell population, including morphology, count, and spatial distribution (see Figure \ref{fig:sirna-batch-image-matrix}). The set of consistent, observable characteristics associated with each siRNA is called its \emph{phenotype}. Note that the phenotype of an siRNA is sometimes visually distinct, but more often its visual characteristics are subtle and hard to detect by eye (see Figure \ref{fig:siRNApheno}).

\subsection{Experiment design}
\label{sec:expdesign}
Of the 308 usable wells on each plate, one is left untreated to provide a negative control (labeled EMPTY), and another 30 wells receive a unique siRNA from a positive control set of 30 siRNA. The remaining 277 wells receive a unique siRNA from a treatment set of 1,108 siRNA. Therefore, each 4-plate experiment contains 1,138 unique siRNA perturbations, where the positive and negative controls appear once on each plate, and the 1,108 treatments appear once in each 4-plate experiment. The location of each siRNA is randomized per experiment and plate, though for operational reasons, the 1,108 treatment siRNA are divided into four groups of 277 that always appear together on a plate. Note that some wells do not receive their intended siRNA (and are thus labeled EMPTY) due to detected operational errors, while images of occasional other wells are removed from the dataset due to detected poor image quality.

\subsection{Image resolution}
Images were acquired at a spatial resolution of $2048\times2048$ and 16 bits per pixel (bpp) per channel, downsampled to $1024\times1024$ at 8bpp, and cropped to the center $512\times512$ field of view. RxRx1 contains two non-overlapping $512\times512$ fields of view per well. Of the possible 125,564 total images (51 experiments $\times$ 4 plates/experiment $\times$ 308 wells/plate $\times$ 2 images/well), 154 images were excluded for failing quality filters, resulting in a total of 125,510 6-channel images in the dataset.

\subsection{Cell types}
The 51 experiments are distributed across four cell types: 24 in HUVEC, 11 in RPE, 11 in HepG2, and 5 in U2OS. Figure \ref{fig:siRNApheno} shows the phenotype of a single siRNA in the four different cell types. Each of the 51 experiments was run in a different batch, resulting in images that exhibit distinct batch effects. It is this feature of the dataset that makes it particularly suited for studying batch effects and methods for correcting them.

\subsection{Metadata}
The following metadata is provided for each image in RxRx1: cell type, experiment id, plate id, well location, and treatment class (1,138 siRNA classes plus one untreated class).


\section{Evaluation task}
We propose the following invariant risk minimization task for evaluating the effectiveness of batch effect correction methods: learn to classify the genetic perturbation present in each image in a set of experimental batches held out from a training set. In order for a classifier to perform well on this task, it must be able to robustly identify the visual characteristics associated with each genetic perturbation against a background of latent, technical variations associated with each experimental batch.

\subsection{Batch-separated vs batch-stratified splits}
\label{sec:splits}
In order to appropriately evaluate such classifiers, we propose two strategies for splitting the data into training and test sets as well as two specific instantiations of these splits. The first, called the \emph{batch-separated} split, assigns 33 of the 51 experiments (16 HUVEC, 7 RPE, 7 HepG2, 3 U2OS) to the training set, and the remaining 18 (8 HUVEC, 5 RPE, 5 HepG2, 2 U2OS) to the test set. In this way, the experimental batches that make up the test set are different from those in the training set, which allows for assessing out-of-domain generalization. Note that this split is naturally stratified with respect to treatment class (see Section \ref{sec:expdesign}). The second split, called the \emph{batch-stratified} split, stratifies the data by both treatment class and experimental batch. The size of the training and test sets are made roughly the same as in the batch-separated split. In the batch-stratified split, the training and test sets contain images from all experimental batches, making the classification task easier to learn since no experimental batch is out-of-domain. As a result, accuracy on the batch-stratified split provides an upper bound for accuracy on the batch-separated split, and we will use both of these numbers when evaluating experimental batch correction methods. Splits can be downloaded at \url{https://rxrx.ai}.

\subsection{Evaluation metrics}
\label{sec:evaluation-metrics}
With the batch-separated and batch-stratified splits defined, we now propose three evaluation metrics for assessing the effectiveness of experimental batch correction methods.

\subsubsection{Perturbation classification accuracy}
This metric is the average perturbation class classification accuracy (including controls and untreated as classes) on the test set when using the batch-separated split. It is useful as an overall measure of the goodness of the batch effect correction method since the test set contains experimental batches not seen during training and the training and test sets are stratified by siRNA classes.

\subsubsection{Batch generalization}
To define a metric that measures generalization to new experimental batches, we calculate perturbation classification accuracy using both the batch-separated and batch-stratified splits, and then measure the difference between these accuracies as follows:
$$\textit{Generalization} = \frac{ \textit{SeparatedPertAcc}}{\textit{StratifiedPertAcc}}$$
where $\textit{SeparatedPertAcc}$ is perturbation classification accuracy on the test set of the batch-separated split (after training on the batch-separated training set), and $\textit{StratifiedPertAcc}$ is perturbation classification accuracy on the test set of the batch-stratified split (after training on batch-stratified training set). A generalization of 100\% means that perturbation classification accuracy on both splits is the same, \ie, the experimental batch correction method has learned to classify perturbations in unseen experimental batches as well as it has learned to classify perturbations in seen experiment batches.

\subsubsection{Batch classification accuracy}
\label{sec:batchclassacc}
To measure whether image embeddings are batch-invariant, we train a separate head atop the embeddings to classify which batch each example is from. We use the batch-stratified split since the training data must contain examples from all batches. Except for later gradient reversal experiments, we do not backpropagate this separate head's loss to the network trunk, so it serves as a simple probe to measure how much batch information is present in the embeddings. We generally want embeddings to be batch-invariant, \ie batch classification accuracy at chance levels of 1/51 $\approx$ 1.96\%.

\begin{figure*}[!t]
    \centering
    \includegraphics[width = 0.99\textwidth]{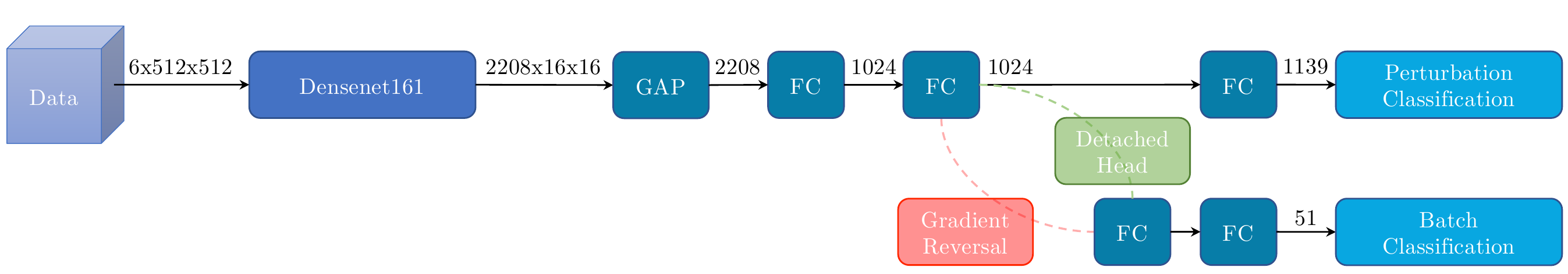}
    \caption{
        A diagram of our models.
        6-channel images are fed to the backbone (DenseNet161 \cite{huang2017densely}).
        Feature maps from the backbone are pooled by global average pooling and then mapped by two fully connected layers, which follow batch normalization and ReLU layers to obtain a 1024-dimensional image embedding.
        The embedding layer is connected to two parallel branches -- one for perturbation classification and the other for experimental batch classification.
        The experimental batch classification branch is either detached (for baseline and AdaBN model) or gradient reversed (for gradient reversal model).
        For both classification targets, we use cross-entropy loss.
    }
    \label{fig:arch}
\end{figure*}

\section{Experimental batch correction methods}
\label{sec:evaluation-methods}
In this section, we describe the methods for experimental batch correction that will be evaluated in this paper using the metrics defined in Section \ref{sec:evaluation-metrics}. 

\subsection{Baseline}
The baseline method is a standard convolutional classifier architecture (see Figure \ref{fig:arch}). A detached batch classification head is added to calculate experimental batch accuracy without backpropagating experimental batch classification error into the rest of the network. For data augmentations, we use horizontal and vertical flips, 90-degree rotations, and CutMix \cite{yun2019cutmix}. We train for 100 epochs, using a cosine learning rate schedule with a 5 epoch linear warmup and learning rate of 0.1024, an SGD optimizer with 0.9 momentum, and a batch size of 512 distributed across 8 Nvidia A100 GPUs. Before feeding an image into the network, we preprocess the image with a channel-wise \emph{self-standardization}, \ie, we subtract the mean and divide it by the standard deviation of the image's pixel intensities per channel.

\subsection{AdaBN}
Adaptive batch normalization (AdaBN) \cite{li2016revisiting} modifies standard batch normalization \cite{ioffe2015batch} layers to use statistics from individual domain distributions (e.g., from experimental batch distributions in our case) rather than the entire dataset distribution, both during training and at test time. Therefore, during training, it is necessary to sample mini-batches from a single experimental batch at a time. By doing so, the model is able to normalize intermediate features within the context of the experimental batch distribution. The rest of the model is unchanged (see Figure \ref{fig:arch}).

\subsection{Gradient reversal}
Gradient reversal \cite{ganin2016domain} is an adversarial method that changes the sign of the gradient for specific layers in the model, e.g., layers connecting the heads of adversarial losses to the rest of the network. Intuitively, this method updates model weights at the gradient reversal layer in order to increase the adversarial loss, while the rest of the head updates its weights in order to decrease the loss, giving rise to the adversarial nature of this method. We want the model to be invariant to differences in experimental batches, thus to implement this method, we reattach the experimental batch classification head mentioned in Section \ref{sec:batchclassacc} using gradient reversal (see Figure \ref{fig:arch}).

\subsection{AdaBN + gradient reversal}
We also apply adaptive batch normalization and gradient reversal simultaneously in order to evaluate their combined ability to correct experimental batch effects.

\section{Experiments}
In this section, we evaluate the methods described in Section \ref{sec:evaluation-methods} using the evaluation metrics described in Section \ref{sec:evaluation-metrics}.

\subsection{Evaluation metric performance}
\begin{table*}[!htb]
    \centering
    \begin{tabular}{@{\hskip 4pt}c@{\hskip 4pt}@{\hskip 4pt}c@{\hskip 4pt}@{\hskip 4pt}c@{\hskip 4pt}@{\hskip 4pt}c@{\hskip 4pt}@{\hskip 4pt}c@{\hskip 4pt}} 
\toprule
\multirow{2}{*}{Method} & Perturbation classification  & Perturbation classification  & Batch & Batch classification    \\
      & accuracy (batch-separated) & accuracy (batch-stratified) & generalization & accuracy \\
\cmidrule{1-5}
      
Baseline & $75.1\% \pm 0.2\%$ & $\textbf{91.1\%} \pm 0.1\%$ & $82.4\%$ & $59.2\% \pm 0.7\%$ \\
Gradient Reversal & $71.2\% \pm 0.4\%$ & $89.1\% \pm 0.1\%$ & $79.9\%$ & $\textbf{\phantom{0}1.8\%} \pm 0.1\%$ \\
AdaBN & $\textbf{87.1\%} \pm 0.2\%$ & $\textbf{91.1\%} \pm 0.1\%$ & $\textbf{95.6\%}$ & $16.4\% \pm 0.3\%$ \\
AdaBN + Grad. Rev. & $86.2\% \pm 0.3\%$ & $90.2\% \pm 0.2\%$ & $\textbf{95.6\%}$ & $\phantom{0}2.3\% \pm 0.1\%$ \\
\bottomrule
    \end{tabular}
    \caption{
        Performance of experimental batch correction methods on the proposed metrics. All models, despite having similar perturbation accuracy on batches seen during training, vary in their ability to generalize to new batches as well as batch classification accuracy. AdaBN improves generalization to new batches significantly, and gradient reversal reduces batch information encoded in embeddings. Using both methods simultaneously yields the benefits of both. For every method, the model was trained 5 times on both batch-separated and batch-stratified splits. For descriptions of the splits, metrics, and methods, see Sections \ref{sec:splits}, \ref{sec:evaluation-metrics}, and  \ref{sec:evaluation-methods}, respectively.
    }
    \label{tab:results}
\end{table*}

The results of the experimental batch correction methods are summarized in Table \ref{tab:results}. The baseline classifier generalizes poorly to new batches, classifying experimental batches about 30x better than random. The AdaBN model improves experimental batch generalization to nearly 96\% while significantly reducing experimental batch classification accuracy ($\sim$8x better than random). Interestingly, gradient reversal does not improve experimental batch generalization but does reduce experimental batch classification accuracy to random chance. Finally, combining AdaBN and gradient reversal yields the benefits of both methods: top experimental batch generalization and near-random experimental batch classification.
\begin{figure*}[!t]
    \centering
    \begin{tabular}{cccc}
        \includegraphics[width=0.28\textwidth]{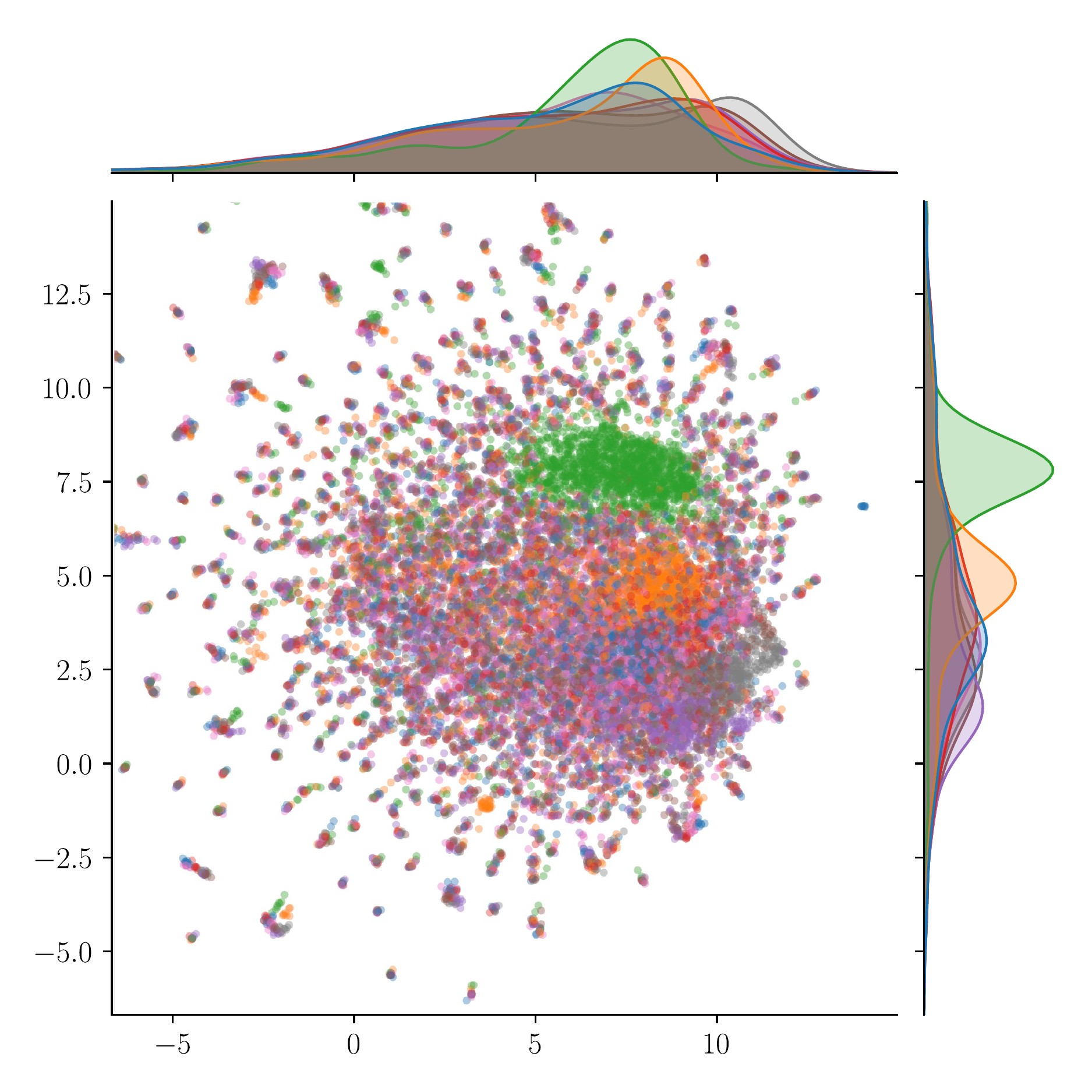} &
        \includegraphics[width=0.28\textwidth]{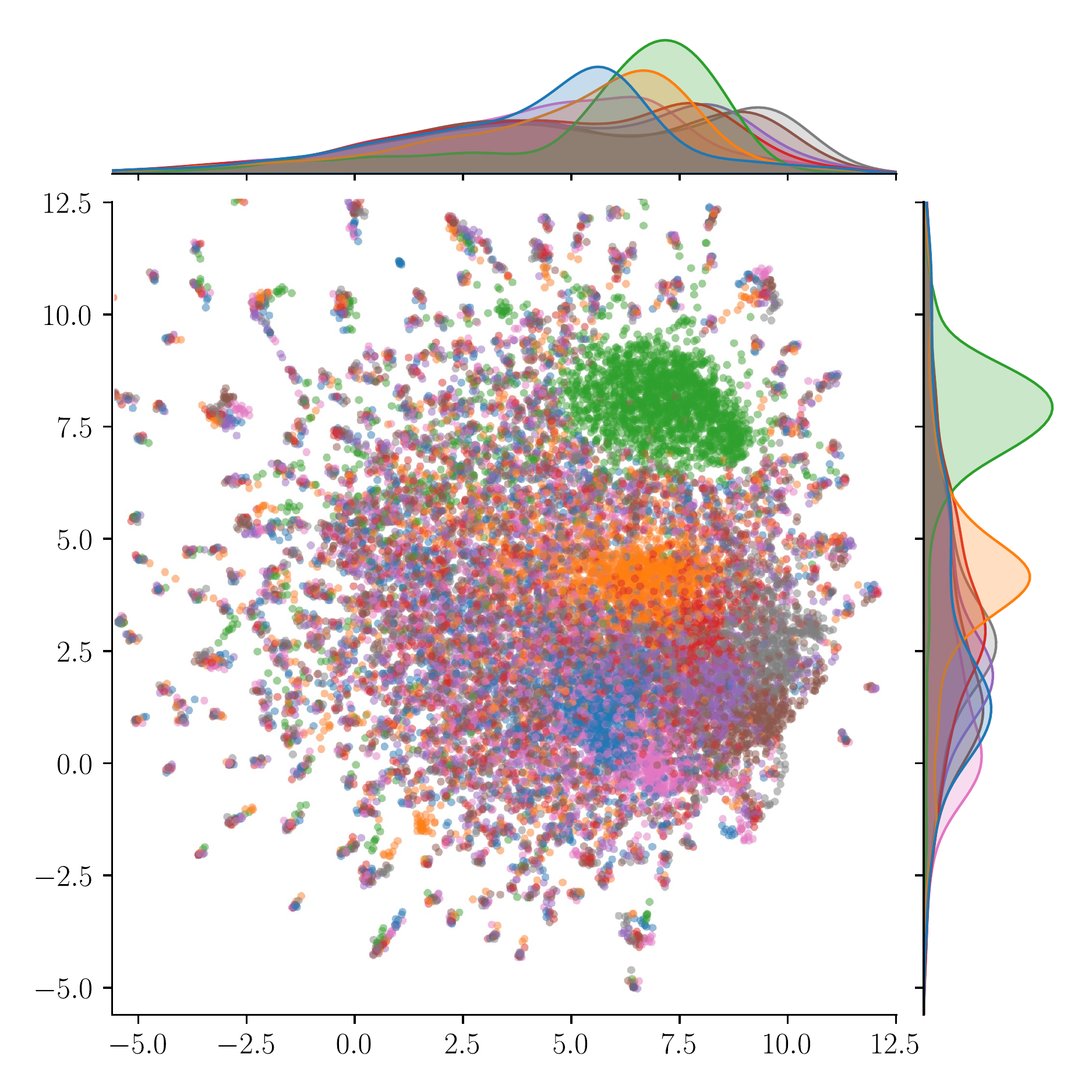} &
        \includegraphics[width=0.28\textwidth]{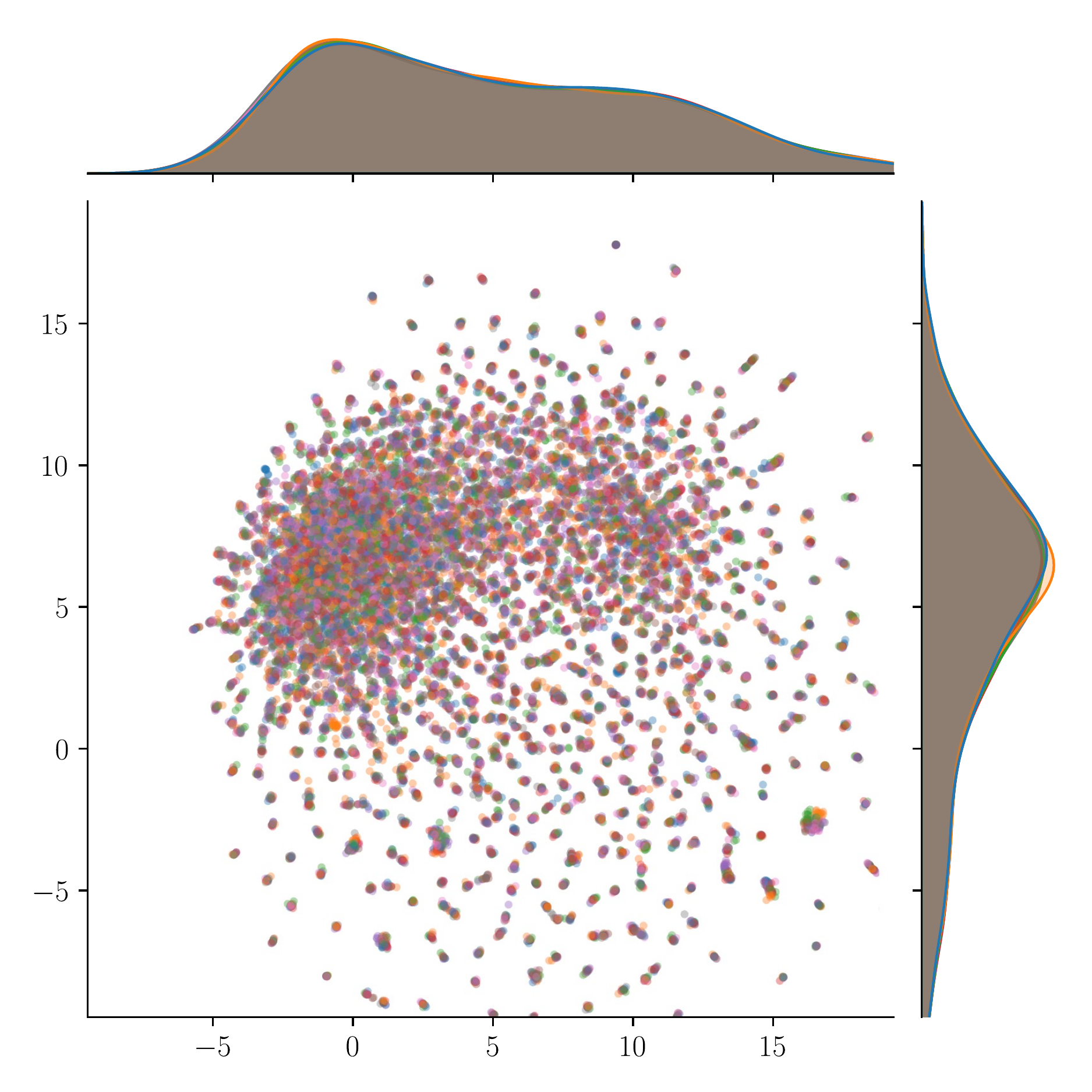} &
        \begin{minipage}{0.085\textwidth}
            \vspace{-4cm}
            \includegraphics[width=\textwidth]{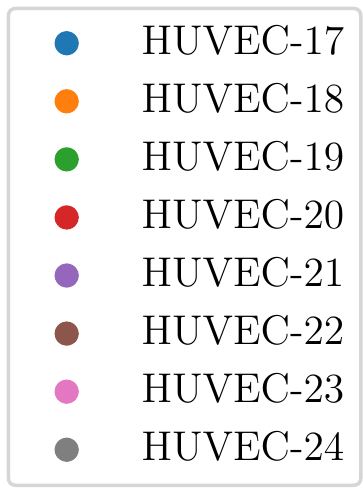}
        \end{minipage}
        \\
        Baseline & Gradient reversal & AdaBN & \\
    \end{tabular}
    \caption{UMAP visualization of embedding spaces for baseline, gradient reversal, and AdaBN methods (AdaBN + gradient reversal UMAP is similar to AdaBN UMAP). Points represent embeddings of individual images in HUVEC experiment batches from the test set and are colored by the experimental batch (other cell types exhibit similar behavior). Note that while gradient reversal is able to reduce experimental batch classification accuracy to random when trained on the batch-stratified split, this behavior does not generalize well to unseen experimental batches. In contrast, AdaBN is far more effective in aligning unseen experiment batches.}
    \label{fig:umap}
\end{figure*}
\subsection{Visualization of embedding space}
In order to gain a better understanding of the information encoded in the embeddings learned by each experimental batch correction method, in Figure \ref{fig:umap} we visualize the learned embedding spaces from our baseline, gradient reversal, and AdaBN methods using UMAP embeddings \cite{mcinnes2018umap}. We note that AdaBN + gradient reversal UMAPs are similar to AdaBN UMAPs. While gradient reversal is able to reduce experimental batch classification accuracy to random when trained on the batch-stratified split, this behavior does not generalize well to experimental batches from unseen experiment batches. In contrast, AdaBN is far more effective in aligning unseen experiment batches since it normalizes intermediate image features with the statistics of the associated experimental batch, rather than the statistics of the training set in standard batch normalization.

\begin{table}[!b]
    \centering
    \begin{tabular}{@{\hskip 5pt}c@{\hskip 5pt}@{\hskip 5pt}c@{\hskip 5pt}@{\hskip 5pt}c@{\hskip 5pt}} 
    \toprule
    Model & Baseline & AdaBN \\
    \cmidrule{1-3}
    Default & 75.2 & 87.1 \\
    \cmidrule{1-3}
    -CutMix & 70.8 & 80.2 \\
    -CutMix +MixUp & 75.6 & 83.9 \\
    backbone=resnet50 & 71.7 & 83.9 \\
    backbone=resnet101 & 71.9 & 84.6 \\
    backbone=densenet121 & 74.3 & 85.9 \\
    \bottomrule
    \end{tabular}\vspace{2mm}
    \caption{
        Perturbation classification accuracy (\%) for various choices of data augmentation methods and convolutional backbones of the baseline and AdaBN methods.
    }
    \label{tab:table2}
\end{table}

\subsection{Effect of data augmentation and backbone choices}
\noindent In Table \ref{tab:table2} we present perturbation classification accuracy results for different choices of data augmentation methods and convolutional backbones. We note how shallower networks have worse performance, and how AdaBN boosts accuracy over the baseline in all scenarios. We also note how using MixUp instead of CutMix augmentation gives better performance for the baseline but not AdaBN.

\subsection{Preservation of embedding similarities}

\begin{figure}[!t]
\centering
\begin{tabular}{c}
  \includegraphics[width=0.76\textwidth]{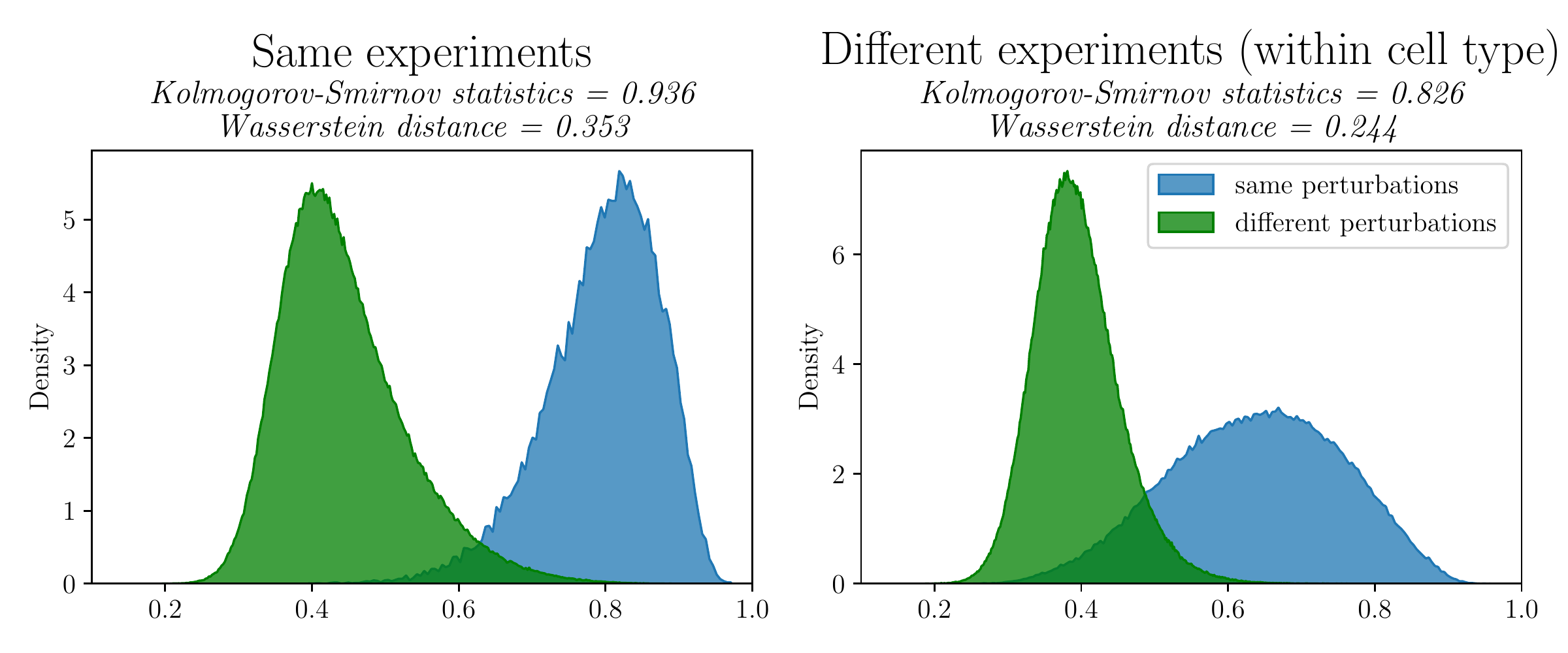}  \\ 
  Baseline\\
  \includegraphics[width=0.76\textwidth]{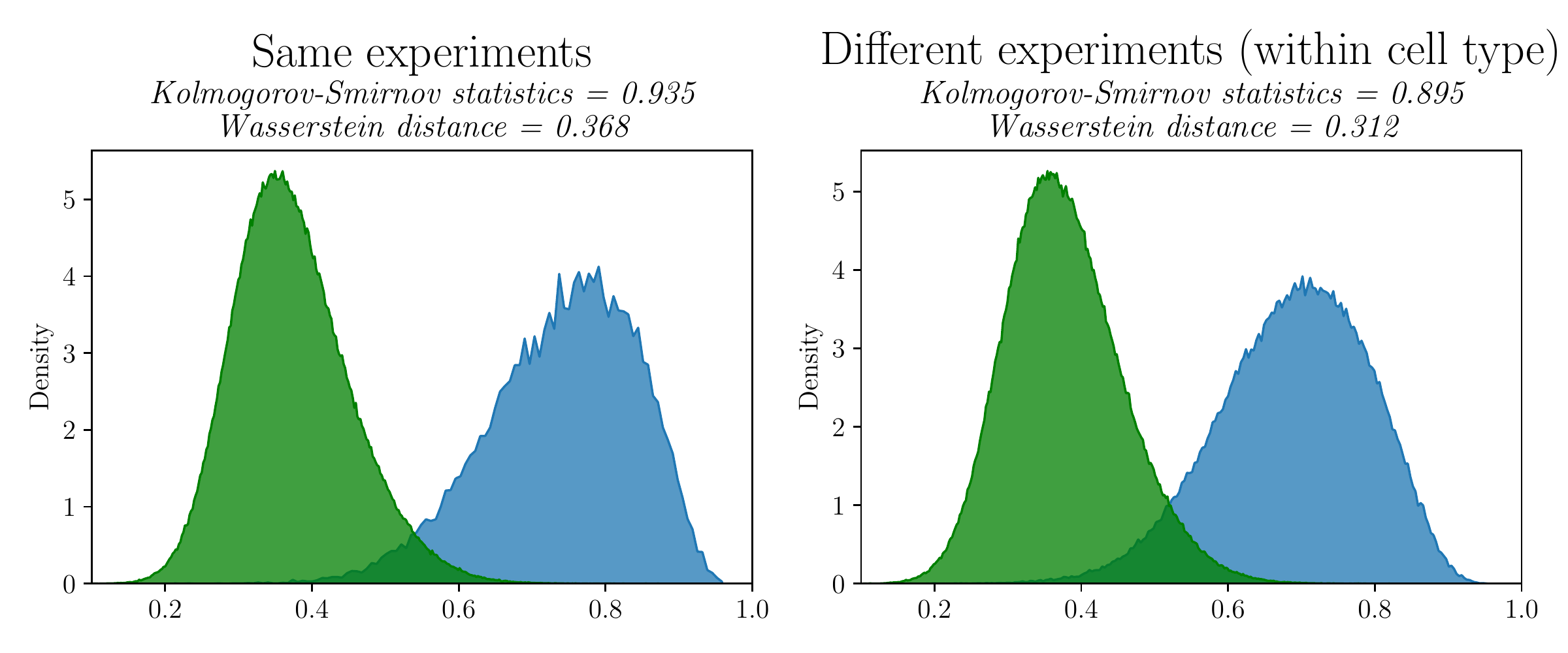}\\
  AdaBN\\
\end{tabular}
    \caption{
        Distributions of cosine similarities between image embeddings for the baseline (\textbf{Top}) and AdaBN methods (\textbf{Bottom}). \textbf{Green}: cosine similarities between different perturbations. \textbf{Blue}: cosine similarities between the same perturbations. \textbf{Left}: cosine similarities between perturbations in the same experimental batches. \textbf{Right}: cosine similarities between perturbations in different experimental batches but the same cell type. Two measures of distributional similarity, KS statistic, and Wasserstein distance, are computed between the two distributions in each plot. Note that the baseline distributions of the same and different perturbation cosine similarities are distinctly different within and across experimental batches, while the AdaBN distributions are very similar, showing that AdaBN preserves geometric relationships between embeddings even across experimental batches.
        Note that cosine similarities are always positive because all values in embeddings are positive as embeddings are obtained by passing features through ReLU in the model.
    }
    \label{fig:perturbation-consistency}
\end{figure}

While the previous section demonstrated that AdaBN is sufficient to align embedding distributions across experimental batches, we also wondered if it would preserve geometric relationships across batches. In order to answer this question, we consider the following distributions of cosine similarities between perturbation embeddings:
\begin{enumerate}
    \item same perturbations in same experimental batches
    \item different perturbations in the same experimental batches
    \item same perturbations in different experimental batches (but same cell type)
    \item different perturbations in different experimental batches (but same cell type)
\end{enumerate}
In Figure \ref{fig:perturbation-consistency}, we compare distributions 1 and 2 with distributions 3 and 4, for both the baseline and AdaBN methods. The similarity of these pairs of distributions to each other would be strong evidence that the experimental batch correction method preserves geometric relationships across experimental batches. Moreover, we calculate two measures of distributional similarity, KS statistic and Wasserstein distance, between each pair of distributions in order to quantify these similarities. As can be seen, the baseline distributions of the same and different perturbation cosine similarities are distinctly different within and across experimental batches, indicating that geometric relationships are not preserved across experimental batches for the baseline method. In contrast, the AdaBN distributions are very similar within and across experiment batches, demonstrating that AdaBN does indeed preserve geometric relationships across experimental batches.





\subsection{Classification accuracy per cell type}

\begin{table}[!t]
    \centering
    \begin{tabular}{@{\hskip 2pt}c@{\hskip 2pt}@{\hskip 2pt}c@{\hskip 2pt}@{\hskip 2pt}c@{\hskip 2pt}@{\hskip 2pt}c@{\hskip 2pt}@{\hskip 2pt}c@{\hskip 2pt}} 
\toprule
Model & HUVEC & RPE & HepG2 & U2OS \\
\cmidrule{1-5}
Baseline & $84.2$ & $79.0$ & $76.2$ & $26.1$ \\[-1.4ex] &\tiny$ \pm 0.2$ & \tiny$ \pm 0.4$ & \tiny$ \pm 0.1$ & \tiny$ \pm 1.5$ \\
Gradient reversal & $83.8$ & $78.1$ & $74.0$ & $24.3$ \\[-1.4ex] &\tiny$ \pm 0.2$ & \tiny$ \pm 0.5$ & \tiny$ \pm 0.6$ & \tiny$ \pm 0.7$ \\
AdaBN & $\textbf{92.1}$ & $87.2$ & $\textbf{86.2}$ & $\textbf{68.2}$ \\[-1.4ex] &\tiny$ \pm 0.2$ & \tiny$ \pm 0.0$ & \tiny$ \pm 0.2$ & \tiny$ \pm 0.1$ \\
AdaBN + gradient reversal & $92.0$ & $\textbf{87.5}$ & $85.6$ & $66.9$ \\[-1.4ex] &\tiny$ \pm 0.0$ & \tiny$ \pm 0.1$ & \tiny$ \pm 0.1$ & \tiny$ \pm 0.3$ \\
\bottomrule
    \end{tabular}
    \vspace{2mm}
    \caption{
        Perturbation classification accuracy (\%) per cell type. Note that increases in perturbation classification accuracy due to AdaBN are larger for more difficult cell types.
    }
    \label{tab:cell-classif}
\end{table}

\begin{table}[!tb]
    \centering
    \begin{tabular}{@{\hskip 2pt}c@{\hskip 2pt}@{\hskip 2pt}c@{\hskip 2pt}@{\hskip 2pt}c@{\hskip 2pt}@{\hskip 2pt}c@{\hskip 2pt}@{\hskip 2pt}c@{\hskip 2pt}} 
\toprule
Model & HUVEC & RPE & HepG2 & U2OS \\
\cmidrule{1-5}
Baseline & $39.5$ & $39.2$ & $31.4$ & $2.8$ \\[-1.4ex] &\tiny$ \pm 0.7$ & \tiny$ \pm 2.0$ & \tiny$ \pm 0.4$ & \tiny$ \pm 1.0$ \\
Gradient reversal & $41.4$ & $38.8$ & $32.3$ & $3.0$ \\[-1.4ex] &\tiny$ \pm 0.5$ & \tiny$ \pm 0.5$ & \tiny$ \pm 0.4$ & \tiny$ \pm 0.3$ \\
AdaBN & $55.1$ & $56.1$ & $\textbf{56.2}$ & $44.0$ \\[-1.4ex] &\tiny$ \pm 1.1$ & \tiny$ \pm 0.4$ & \tiny$ \pm 1.6$ & \tiny$ \pm 0.9$ \\
AdaBN + gradient reversal & $\textbf{55.3}$ & $\textbf{56.7}$ & $55.5$ & $\textbf{44.1}$ \\[-1.4ex] &\tiny$ \pm 2.0$ & \tiny$ \pm 0.9$ & \tiny$ \pm 0.6$ & \tiny$ \pm 1.4$ \\
\bottomrule
    \end{tabular}\vspace{2mm}
    \caption{
        Perturbation classification accuracy (\%) per cell type on simplified training sets containing only 3 experiments of a single cell type. HUVEC, RPE, and HepG2 cell types are easier to learn than U2OS, however, AdaBN significantly improves all classification accuracies, especially U2OS.
    }
    \label{tab:cell-classif-3exp}
\end{table}

Table~\ref{tab:cell-classif} shows perturbation classification accuracy for each of the four cell types. Note that HUVEC accuracies are highest, followed by RPE and HepG2, and finally U2OS. This is in line with the differing proportions of experimental batches per each cell type in the training set. In order to obtain a more fair comparison of per-cell perturbation classification accuracy, we randomly selected 3 experimental batches for each cell type from the original training set to form a new training set. The results are shown in Table~\ref{tab:cell-classif-3exp}. We note that the HUVEC, RPE, and HepG2 cell types are far easier to learn than U2OS; however, AdaBN significantly improves classification accuracies in all cell types, especially U2OS. Comparing the results (for U2OS since training sets were the same in both) in Tables \ref{tab:cell-classif} and \ref{tab:cell-classif-3exp}, we conclude that jointly training a method on all cell types rather than individual cell types greatly improves perturbation classification accuracy.


\begin{figure}[!tb]
    \centering
    \includegraphics[width=0.37\textwidth]{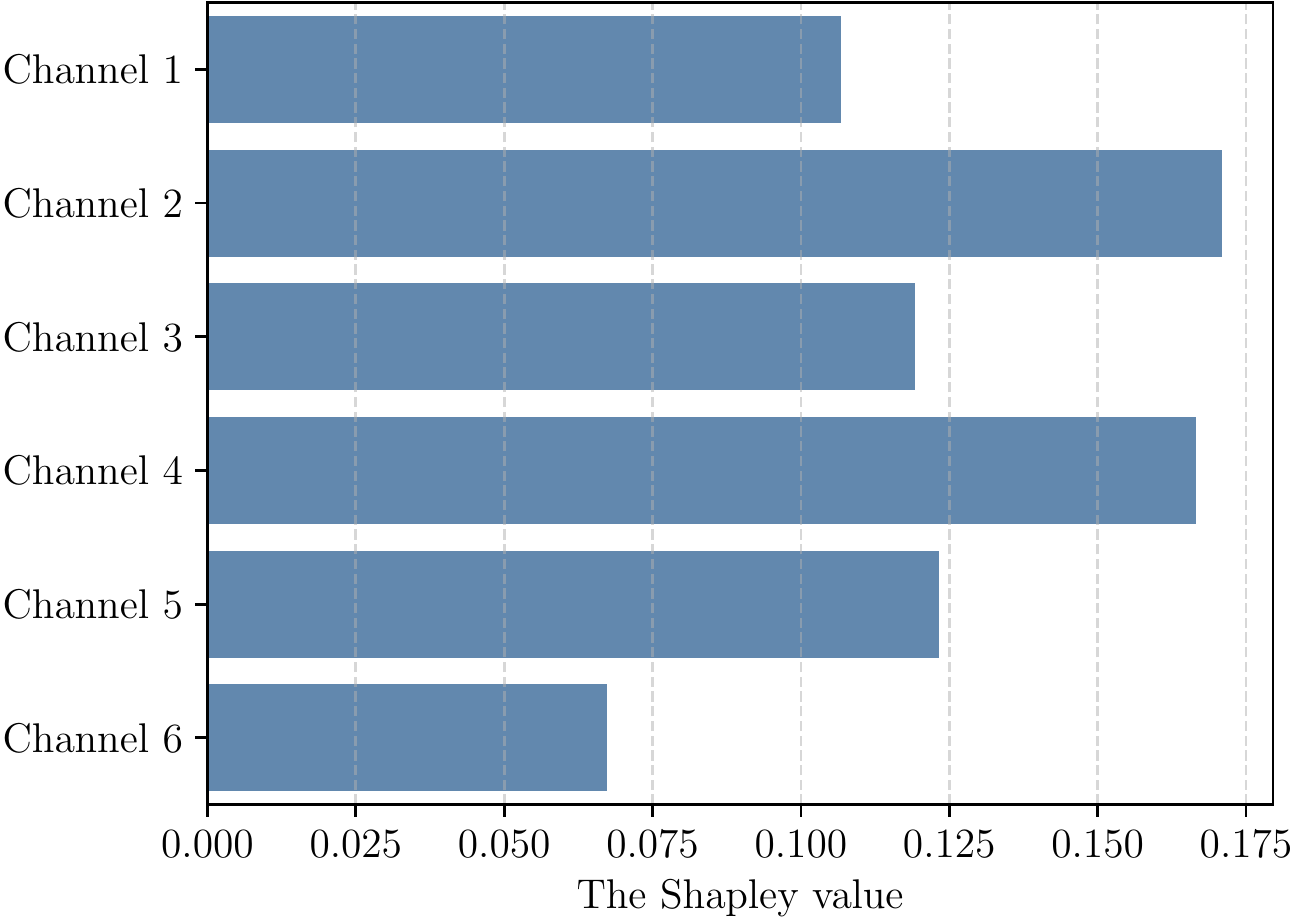}
    \caption{
        The Shapley values for each channel using the baseline method, represent contributions to perturbation classification accuracy. Higher values represent greater importance. Similar Shapley values are observed for each experimental batch correction method.
    }    
    \label{fig:channels-shapley}
\end{figure}

\subsection{Channel importance}

\begin{figure}[!tb]
    \centering
    \includegraphics[width = 0.46\textwidth]{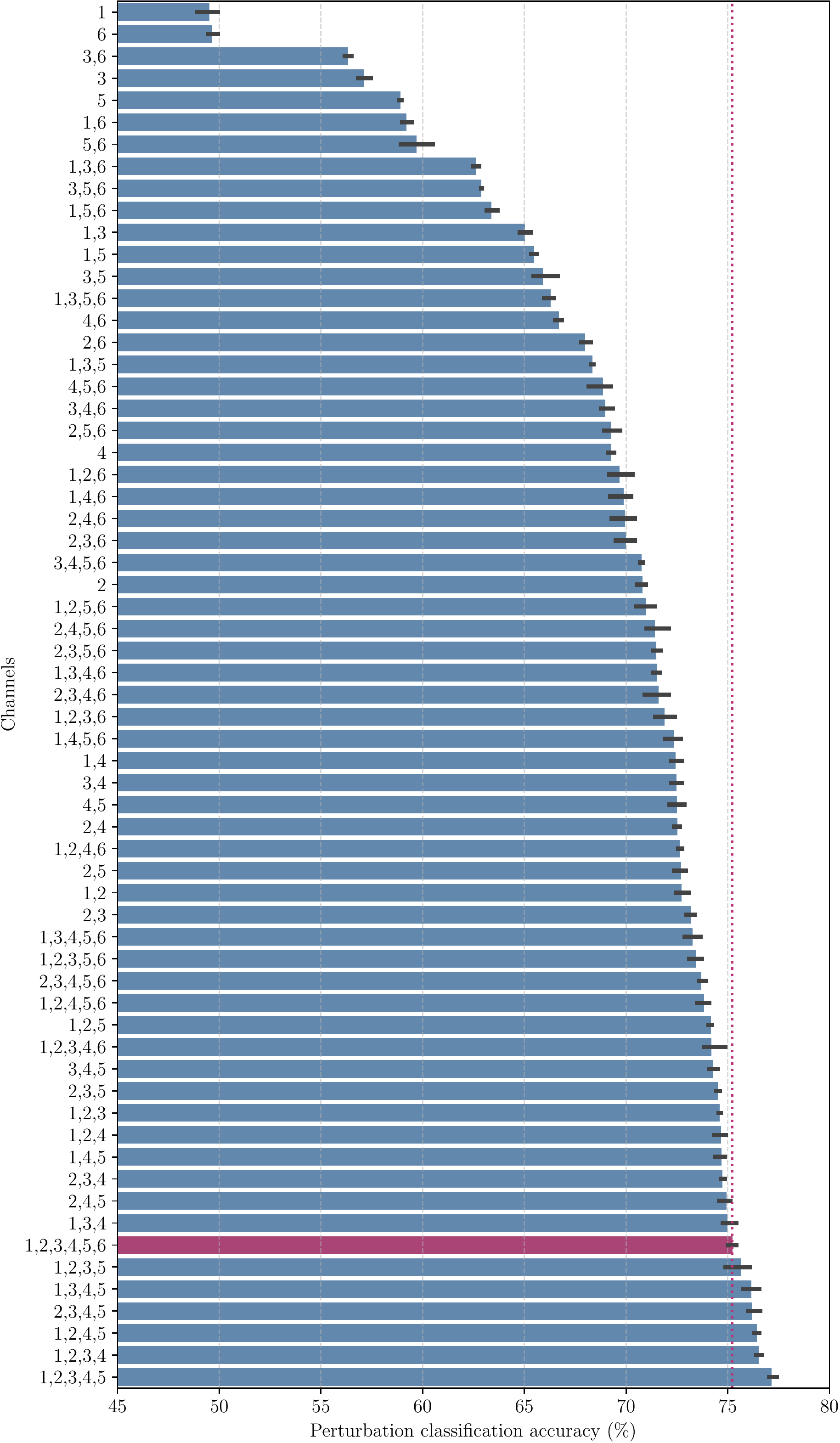}
    \caption{Perturbation classification accuracy of baseline method trained on different subsets of channels.}   
    \label{fig:newFig8}
\end{figure}

In Figure~\ref{fig:channels-shapley}, we study the importance of each channel by plotting its Shapley value \cite{shapley1951notes} for the baseline method. Shapley values measure the relative contributions each channel makes when assigning correct classes to our (batch-separated) test set. Channels 2 and 4 are most important, while Channel 6 is the least important. We note that the large importance of Channels 2 and 4 is likely due to the spectral overlap that their fluorescent stains have with the stains of other channels in Recursion's protocol, specifically Channel 1 (which consequently lowers the relative accuracy contribution of Channel 1). In Figure \ref{fig:newFig8} we show perturbation classification accuracy of the baseline method trained on all non-empty subsets of channels. Interestingly, we observe that the model that uses all 6 channels does not yield the best performance. All channel subsets containing at least 4 channels without Channel 6 surpass the all-channels baseline. Using all but Channel 6 exceeds the baseline by 2 percentage points. We hypothesize that this observation is a result of overfitting the model on RxRx1 with its particular channel statistics, not that models with fewer input channels will in general outperform models with access to all input channels.

\subsection{Image preprocessing}
We tried different image normalization methods for preprocessing the images. In all cases, we calculate per-channel means and standard deviations on different subsets of the dataset and standardize each image with those statistics (\ie, subtract the mean and divide by the standard deviation) before using them as input to the networks. The (batch-separated) perturbation classification accuracies associated with each normalization method are presented in Table~\ref{tab:image-normalization}. Self-standardization (standardization using only the image itself) outperforms other methods by a significant margin. Interestingly, the self-standardization improves perturbation classification accuracy by more than 14 percentage points compared to the common computer vision practice of using global statistics calculated from the entire training set. We hypothesize that this margin is due to the uniform background of RxRx1 images across experimental batches, which contains little biological information but whose size relative to the foreground of cells can change dramatically from perturbation to perturbation and even image to image. Thus image-level statistics are proportional to the cellular content of an image, so that self-standardization normalizes each image to the common scale of the average cell contained within the image.

\begin{table}
    \centering
    \begin{tabular}{ccc}
\toprule
Normalization               & Accuracy \\
\cmidrule{1-2}
All images                      & $60.8 \pm 1.1$ \\
Control images per experiment   & $68.4 \pm 0.5$ \\
All images per experiment       & $68.6 \pm 0.5$ \\
Control images per plate        & $73.4 \pm 0.3$ \\
All images per plate            & $73.4 \pm 0.5$ \\
Self-standardization   & $\textbf{75.1} \pm 0.2$ \\
\bottomrule
    \end{tabular}\vspace{2mm}
    \caption{
        Perturbation classification accuracy (\%) for different image normalization methods on the batch-separated split using the baseline method. Self-standardization, where each channel of a single image is standardized by its own mean and standard deviation, yields the best results.
    }
    \label{tab:image-normalization}
\end{table}

\noindent In Table \ref{tab:table5} we show perturbation classification accuracy for different image normalization methods using our AdaBN method. Similar to Table 4, self-standardization (\ie, per-image statistics) offers the best perturbation classification accuracy.

\begin{table}
    \centering
    \begin{tabular}{ccc}
        \hline
        \toprule
        Normalization               & Accuracy \\
        \cmidrule{1-2}
        All images                      & $78.4$ \\
        Control images per experiment   & $83.6$ \\
        All images per experiment       & $83.7$ \\
        Control images per plate        & $81.7$ \\
        All images per plate            & $82.6$ \\
        Self-standardization    & $\textbf{87.1}$ \\
        \bottomrule
    \end{tabular}\vspace{2mm}
    \caption{
        Perturbation classification accuracy (\%) for different image normalization methods using AdaBN.}
     \label{tab:table5}
\end{table}


\subsection{Training dynamics}
\begin{figure}[!ht]
    \centering
    \includegraphics[width = 0.49\textwidth]{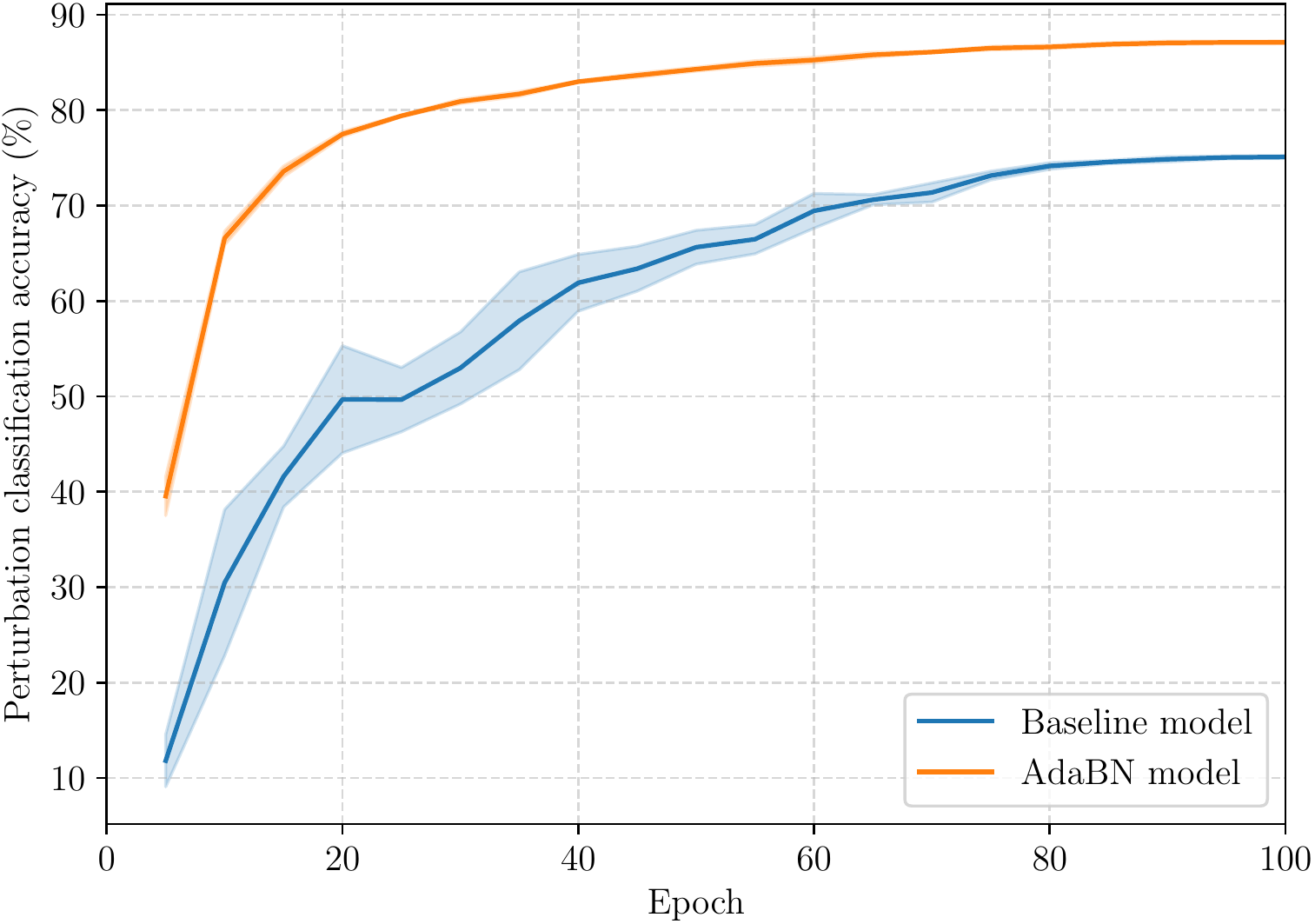}
    \caption{
        Test perturbation classification accuracy during model training (mean and standard deviation from 5 runs). AdaBN model converges faster, and the runs are more consistent with each other.
    }   
    \label{fig:figS1}
\end{figure}

\noindent In Figure \ref{fig:figS1}, we plot test perturbation classification accuracy means and standard deviations during model training (over 5 runs). Our results show that the model architecture with AdaBN converges faster than the baseline. Moreover, AdaBN has a much smaller standard deviation than the baseline, i. e., the each run is more consistent with the others for AdaBN.

\section{Conclusion}
In this paper, we described the \emph{RxRx1} dataset, an image dataset systematically designed to study experimental batch effect correction methods. 
The dataset contains 125,510 6-channel, high-resolution fluorescence microscopy images of human cells under 1,138 genetic perturbations in 51 experimental batches across 4 cell types.
We proposed a task and several metrics to evaluate the performance of different experimental batch correction methods. We demonstrated that while both adaptive batch normalization (AdaBN) \cite{li2016revisiting} and gradient reversal \cite{ganin2016domain} are effective techniques for removing experimental batch information from image embeddings, only AdaBN was effective in generalizing to unseen experimental batches, due to the manner in which it normalizes all intermediate feature maps using statistics from the corresponding experimental batch. We also demonstrated the importance of each image channel in this task, and the value of self-standardization as an image preprocessing step. We hope that the introduction of the RxRx1 dataset will encourage further research into the complex problem of correcting experimental batch effect, as well as other issues that arise in the analysis of high-throughput screening data.

\subsection{Future directions}
There are several methodologies for extracting features from microscopy imaging screens, including manual feature extraction (e.g., using CellProfiler) \cite{singh2014pipeline,ljosa2013comparison}, leveraging pre-trained deep learning models \cite{ando2017improving,pawlowski2016automating}, and training deep learning models on microscopy images directly \cite{godinez2017multi}. As both AdaBN and gradient reversal are deep learning methodologies, it is not possible to directly apply these methods to traditional feature extraction pipelines, yet an appropriate comparison would be useful to understand the benefit of end-to-end feature training.

Our approach relies on weakly supervised learning \cite{caicedo2018weakly,moshkov2022learning} since we train models to predict the experimental perturbation in each well, without validating that each treatment induces a unique visual phenotype (N.B.: such validation is likely impossible). This means that there might be multiple perturbations which either do not perturb the cellular morphology or perturb it in similar ways to other perturbations, yet the perturbation classification task would reward distinguishing them. This would encourage reliance on spurious features or correlation, which inhibits learning image representations that capture meaningful morphological features and generalizing out of batch. Recently, self-supervised methods have been shown to match the performance of supervised models on natural image computer vision tasks \cite{chen2020simple,caron2021emerging,assran2022masked}. Applying such training techniques for microscopy screening data \cite{perakis2021contrastive,cross2022self} represents a potentially fruitful direction for this work.

Finally, we acknowledge that the proposed perturbation classification task groups any morphological variation not associated with a common perturbation under the umbrella term \textit{experimental batch effect}, which is usually reserved for technical effects only. One could imagine improving the task in a way that would not penalize intrinsic morphological features, like those associated with cell type differences, even if they are not associated with variations amongst perturbations. Such a task would promote the development of more effective experimental batch correction methods that better disentangle biological and technical causal factors, and we hope to provide such an update to this work in the future.


\bibliographystyle{unsrt}  
\bibliography{main}

\begin{thebibliography}{10}

\bibitem{echeverri2006high}
Christophe~J Echeverri and Norbert Perrimon.
\newblock High-throughput rnai screening in cultured cells: a user's guide.
\newblock {\em Nature Reviews Genetics}, 7(5):373--384, 2006.

\bibitem{zhou2014high}
Yuexin Zhou, Shiyou Zhu, Changzu Cai, Pengfei Yuan, Chunmei Li, Yanyi Huang,
  and Wensheng Wei.
\newblock High-throughput screening of a crispr/cas9 library for functional
  genomics in human cells.
\newblock {\em Nature}, 509(7501):487--491, 2014.

\bibitem{broach1996high}
James~R Broach and Jeremy Thorner.
\newblock High-throughput screening for drug discovery.
\newblock {\em Nature}, 384(6604 Suppl):14--16, 1996.

\bibitem{macarron2011impact}
Ricardo Macarron, Martyn~N Banks, Dejan Bojanic, David~J Burns, Dragan~A
  Cirovic, Tina Garyantes, Darren~VS Green, Robert~P Hertzberg, William~P
  Janzen, Jeff~W Paslay, et~al.
\newblock Impact of high-throughput screening in biomedical research.
\newblock {\em Nature reviews Drug discovery}, 10(3):188--195, 2011.

\bibitem{swinney2011were}
David~C Swinney and Jason Anthony.
\newblock How were new medicines discovered?
\newblock {\em Nature reviews Drug discovery}, 10(7):507--519, 2011.

\bibitem{boutros2015microscopy}
Michael Boutros, Florian Heigwer, and Christina Laufer.
\newblock Microscopy-based high-content screening.
\newblock {\em Cell}, 163(6):1314--1325, 2015.

\bibitem{scannell2012diagnosing}
Jack~W Scannell, Alex Blanckley, Helen Boldon, and Brian Warrington.
\newblock Diagnosing the decline in pharmaceutical r\&d efficiency.
\newblock {\em Nature reviews Drug discovery}, 11(3):191--200, 2012.

\bibitem{dimasi2016innovation}
Joseph~A DiMasi, Henry~G Grabowski, and Ronald~W Hansen.
\newblock Innovation in the pharmaceutical industry: new estimates of r\&d
  costs.
\newblock {\em Journal of health economics}, 47:20--33, 2016.

\bibitem{leek2010tackling}
Jeffrey~T Leek, Robert~B Scharpf, H{\'e}ctor~Corrada Bravo, David Simcha,
  Benjamin Langmead, W~Evan Johnson, Donald Geman, Keith Baggerly, and Rafael~A
  Irizarry.
\newblock Tackling the widespread and critical impact of batch effects in
  high-throughput data.
\newblock {\em Nature Reviews Genetics}, 11(10):733--739, 2010.

\bibitem{parker2012practical}
Hilary~S Parker and Jeffrey~T Leek.
\newblock The practical effect of batch on genomic prediction.
\newblock {\em Statistical applications in genetics and molecular biology},
  11(3), 2012.

\bibitem{soneson2014batch}
Charlotte Soneson, Sarah Gerster, and Mauro Delorenzi.
\newblock Batch effect confounding leads to strong bias in performance
  estimates obtained by cross-validation.
\newblock {\em PloS one}, 9(6):e100335, 2014.

\bibitem{nygaard2016methods}
Vegard Nygaard, Einar~Andreas R{\o}dland, and Eivind Hovig.
\newblock Methods that remove batch effects while retaining group differences
  may lead to exaggerated confidence in downstream analyses.
\newblock {\em Biostatistics}, 17(1):29--39, 2016.

\bibitem{korsunsky2019fast}
Ilya Korsunsky, Nghia Millard, Jean Fan, Kamil Slowikowski, Fan Zhang, Kevin
  Wei, Yuriy Baglaenko, Michael Brenner, Po-ru Loh, and Soumya Raychaudhuri.
\newblock Fast, sensitive and accurate integration of single-cell data with
  harmony.
\newblock {\em Nature methods}, 16(12):1289--1296, 2019.

\bibitem{hie2019efficient}
Brian Hie, Bryan Bryson, and Bonnie Berger.
\newblock Efficient integration of heterogeneous single-cell transcriptomes
  using scanorama.
\newblock {\em Nature biotechnology}, 37(6):685--691, 2019.

\bibitem{lopez2018deep}
Romain Lopez, Jeffrey Regier, Michael~B Cole, Michael~I Jordan, and Nir Yosef.
\newblock Deep generative modeling for single-cell transcriptomics.
\newblock {\em Nature methods}, 15(12):1053--1058, 2018.

\bibitem{li2020deep}
Xiangjie Li, Kui Wang, Yafei Lyu, Huize Pan, Jingxiao Zhang, Dwight Stambolian,
  Katalin Susztak, Muredach~P Reilly, Gang Hu, and Mingyao Li.
\newblock Deep learning enables accurate clustering with batch effect removal
  in single-cell rna-seq analysis.
\newblock {\em Nature communications}, 11(1):1--14, 2020.

\bibitem{Lotfollahi2020}
Mohammad Lotfollahi, Mohsen Naghipourfar, Fabian~J Theis, and F~Alexander Wolf.
\newblock {Conditional out-of-distribution generation for unpaired data using
  transfer VAE}.
\newblock {\em Bioinformatics}, 36(Supplement 2):610--617, 12 2020.

\bibitem{haghverdi2018batch}
Laleh Haghverdi, Aaron~TL Lun, Michael~D Morgan, and John~C Marioni.
\newblock Batch effects in single-cell rna-sequencing data are corrected by
  matching mutual nearest neighbors.
\newblock {\em Nature biotechnology}, 36(5):421--427, 2018.

\bibitem{goh2017batch}
Wilson Wen~Bin Goh, Wei Wang, and Limsoon Wong.
\newblock Why batch effects matter in omics data, and how to avoid them.
\newblock {\em Trends in biotechnology}, 35(6):498--507, 2017.

\bibitem{shaham2017removal}
Uri Shaham, Kelly~P Stanton, Jun Zhao, Huamin Li, Khadir Raddassi, Ruth
  Montgomery, and Yuval Kluger.
\newblock Removal of batch effects using distribution-matching residual
  networks.
\newblock {\em Bioinformatics}, 33(16):2539--2546, 2017.

\bibitem{arjovsky2019invariant}
Martin Arjovsky, L{\'e}on Bottou, Ishaan Gulrajani, and David Lopez-Paz.
\newblock Invariant risk minimization.
\newblock {\em arXiv preprint arXiv:1907.02893}, 2019.

\bibitem{5206848}
Jia Deng, Wei Dong, Richard Socher, Li-Jia Li, Kai Li, and Li~Fei-Fei.
\newblock Imagenet: A large-scale hierarchical image database.
\newblock In {\em 2009 IEEE Conference on Computer Vision and Pattern
  Recognition}, pages 248--255, 2009.

\bibitem{ljosa2012annotated}
Vebjorn Ljosa, Katherine~L Sokolnicki, and Anne~E Carpenter.
\newblock Annotated high-throughput microscopy image sets for validation.
\newblock {\em Nature methods}, 9(7):637--637, 2012.

\bibitem{angermueller2016deep}
Christof Angermueller, Tanel P{\"a}rnamaa, Leopold Parts, and Oliver Stegle.
\newblock Deep learning for computational biology.
\newblock {\em Molecular systems biology}, 12(7):878, 2016.

\bibitem{kraus2016classifying}
Oren~Z Kraus, Jimmy~Lei Ba, and Brendan~J Frey.
\newblock Classifying and segmenting microscopy images with deep multiple
  instance learning.
\newblock {\em Bioinformatics}, 32(12):i52--i59, 2016.

\bibitem{caicedo2017data}
Juan~C Caicedo, Sam Cooper, Florian Heigwer, Scott Warchal, Peng Qiu, Csaba
  Molnar, Aliaksei~S Vasilevich, Joseph~D Barry, Harmanjit~Singh Bansal, Oren
  Kraus, et~al.
\newblock Data-analysis strategies for image-based cell profiling.
\newblock {\em Nature methods}, 14(9):849--863, 2017.

\bibitem{kraus2017automated}
Oren~Z Kraus, Ben~T Grys, Jimmy Ba, Yolanda Chong, Brendan~J Frey, Charles
  Boone, and Brenda~J Andrews.
\newblock Automated analysis of high-content microscopy data with deep
  learning.
\newblock {\em Molecular systems biology}, 13(4):924, 2017.

\bibitem{ando2017improving}
D~Michael Ando, Cory~Y McLean, and Marc Berndl.
\newblock Improving phenotypic measurements in high-content imaging screens.
\newblock {\em BioRxiv}, page 161422, 2017.

\bibitem{chen2018rise}
Hongming Chen, Ola Engkvist, Yinhai Wang, Marcus Olivecrona, and Thomas
  Blaschke.
\newblock The rise of deep learning in drug discovery.
\newblock {\em Drug discovery today}, 23(6):1241--1250, 2018.

\bibitem{davis2007isolation}
Jaeger Davis, Steve~P Crampton, and Christopher C~W Hughes.
\newblock Isolation of human umbilical vein endothelial cells ({HUVEC}).
\newblock {\em JoVE}, (3):e183, April 2007.

\bibitem{yang2021functions}
Song Yang, Jun Zhou, and Dengwen Li.
\newblock Functions and diseases of the retinal pigment epithelium.
\newblock {\em Frontiers in Pharmacology}, page 1976, 2021.

\bibitem{donato2015culture}
Mar{\'\i}a~Teresa Donato, Laia Tolosa, and Mar{\'\i}a~Jos{\'e}
  G{\'o}mez-Lech{\'o}n.
\newblock Culture and functional characterization of human hepatoma hepg2
  cells.
\newblock In {\em Protocols in In Vitro Hepatocyte Research}, pages 77--93.
  Springer, 2015.

\bibitem{niforou2008proteome}
Katerina~N Niforou, Athanasios~K Anagnostopoulos, Konstantinos Vougas, Christos
  Kittas, Vassilis~G Gorgoulis, and George~T Tsangaris.
\newblock The proteome profile of the human osteosarcoma u2os cell line.
\newblock {\em Cancer genomics \& proteomics}, 5(1):63--77, 2008.

\bibitem{bray2016cell}
Mark-Anthony Bray, Shantanu Singh, Han Han, Chadwick~T Davis, Blake Borgeson,
  Cathy Hartland, Maria Kost-Alimova, Sigrun~M Gustafsdottir, Christopher~C
  Gibson, and Anne~E Carpenter.
\newblock Cell painting, a high-content image-based assay for morphological
  profiling using multiplexed fluorescent dyes.
\newblock {\em Nature protocols}, 11(9):1757--1774, 2016.

\bibitem{tuschl2001rna}
Thomas Tuschl.
\newblock Rna interference and small interfering rnas.
\newblock {\em Chembiochem}, 2(4):239--245, 2001.

\bibitem{huang2017densely}
Gao Huang, Zhuang Liu, Laurens Van Der~Maaten, and Kilian~Q Weinberger.
\newblock Densely connected convolutional networks.
\newblock In {\em Proceedings of the IEEE conference on computer vision and
  pattern recognition}, pages 4700--4708, 2017.

\bibitem{yun2019cutmix}
Sangdoo Yun, Dongyoon Han, Seong~Joon Oh, Sanghyuk Chun, Junsuk Choe, and
  Youngjoon Yoo.
\newblock Cutmix: Regularization strategy to train strong classifiers with
  localizable features.
\newblock In {\em Proceedings of the IEEE/CVF international conference on
  computer vision}, pages 6023--6032, 2019.

\bibitem{li2016revisiting}
Yanghao Li, Naiyan Wang, Jianping Shi, Jiaying Liu, and Xiaodi Hou.
\newblock Revisiting batch normalization for practical domain adaptation.
\newblock {\em arXiv preprint arXiv:1603.04779}, 2016.

\bibitem{ioffe2015batch}
Sergey Ioffe and Christian Szegedy.
\newblock Batch normalization: Accelerating deep network training by reducing
  internal covariate shift.
\newblock In {\em International conference on machine learning}, pages
  448--456. PMLR, 2015.

\bibitem{ganin2016domain}
Yaroslav Ganin, Evgeniya Ustinova, Hana Ajakan, Pascal Germain, Hugo
  Larochelle, Fran{\c{c}}ois Laviolette, Mario Marchand, and Victor Lempitsky.
\newblock Domain-adversarial training of neural networks.
\newblock {\em The journal of machine learning research}, 17(1):2096--2030,
  2016.

\bibitem{mcinnes2018umap}
Leland McInnes, John Healy, and James Melville.
\newblock Umap: Uniform manifold approximation and projection for dimension
  reduction.
\newblock {\em arXiv preprint arXiv:1802.03426}, 2018.

\bibitem{shapley1951notes}
Lloyd~S Shapley.
\newblock Notes on the n-person game—ii: The value of an n-person
  game.(1951).
\newblock {\em Lloyd S Shapley}, 1951.

\bibitem{singh2014pipeline}
Shantanu Singh, M-A Bray, TR~Jones, and AE~Carpenter.
\newblock Pipeline for illumination correction of images for high-throughput
  microscopy.
\newblock {\em Journal of microscopy}, 256(3):231--236, 2014.

\bibitem{ljosa2013comparison}
Vebjorn Ljosa, Peter~D Caie, Rob Ter~Horst, Katherine~L Sokolnicki, Emma~L
  Jenkins, Sandeep Daya, Mark~E Roberts, Thouis~R Jones, Shantanu Singh,
  Auguste Genovesio, et~al.
\newblock Comparison of methods for image-based profiling of cellular
  morphological responses to small-molecule treatment.
\newblock {\em Journal of biomolecular screening}, 18(10):1321--1329, 2013.

\bibitem{pawlowski2016automating}
Nick Pawlowski, Juan~C Caicedo, Shantanu Singh, Anne~E Carpenter, and Amos
  Storkey.
\newblock Automating morphological profiling with generic deep convolutional
  networks.
\newblock {\em BioRxiv}, page 085118, 2016.

\bibitem{godinez2017multi}
William~J Godinez, Imtiaz Hossain, Stanley~E Lazic, John~W Davies, and Xian
  Zhang.
\newblock A multi-scale convolutional neural network for phenotyping
  high-content cellular images.
\newblock {\em Bioinformatics}, 33(13):2010--2019, 2017.

\bibitem{caicedo2018weakly}
Juan~C Caicedo, Claire McQuin, Allen Goodman, Shantanu Singh, and Anne~E
  Carpenter.
\newblock Weakly supervised learning of single-cell feature embeddings.
\newblock In {\em Proceedings of the IEEE Conference on Computer Vision and
  Pattern Recognition}, pages 9309--9318, 2018.

\bibitem{moshkov2022learning}
Nikita Moshkov, Michael Bornholdt, Santiago Benoit, Claire McQuin, Matthew
  Smith, Allen Goodman, Rebecca Senft, Yu~Han, Mehrtash Babadi, Peter Horvath,
  et~al.
\newblock Learning representations for image-based profiling of perturbations.
\newblock {\em bioRxiv}, 2022.

\bibitem{chen2020simple}
Ting Chen, Simon Kornblith, Mohammad Norouzi, and Geoffrey Hinton.
\newblock A simple framework for contrastive learning of visual
  representations.
\newblock In {\em International conference on machine learning}, pages
  1597--1607. PMLR, 2020.

\bibitem{caron2021emerging}
Mathilde Caron, Hugo Touvron, Ishan Misra, Herv{\'e} J{\'e}gou, Julien Mairal,
  Piotr Bojanowski, and Armand Joulin.
\newblock Emerging properties in self-supervised vision transformers.
\newblock In {\em Proceedings of the IEEE/CVF International Conference on
  Computer Vision}, pages 9650--9660, 2021.

\bibitem{assran2022masked}
Mahmoud Assran, Mathilde Caron, Ishan Misra, Piotr Bojanowski, Florian Bordes,
  Pascal Vincent, Armand Joulin, Michael Rabbat, and Nicolas Ballas.
\newblock Masked siamese networks for label-efficient learning.
\newblock {\em arXiv preprint arXiv:2204.07141}, 2022.

\bibitem{perakis2021contrastive}
Alexis Perakis, Ali Gorji, Samriddhi Jain, Krishna Chaitanya, Simone Rizza, and
  Ender Konukoglu.
\newblock Contrastive learning of single-cell phenotypic representations for
  treatment classification.
\newblock In {\em International Workshop on Machine Learning in Medical
  Imaging}, pages 565--575. Springer, 2021.

\bibitem{cross2022self}
Jan~Oscar Cross-Zamirski, Guy Williams, Elizabeth Mouchet, Carola-Bibiane
  Sch{\"o}nlieb, Riku Turkki, and Yinhai Wang.
\newblock Self-supervised learning of phenotypic representations from cell
  images with weak labels.
\newblock {\em arXiv preprint arXiv:2209.07819}, 2022.

\end{thebibliography}

\end{document}